\documentclass[10pt,twocolumn,letterpaper]{article}

\usepackage{cvpr}
\usepackage[accsupp]{axessibility}
\usepackage{booktabs}
\usepackage{multirow}
\usepackage{graphicx}
\usepackage{ulem}

\usepackage{tcolorbox}
\usepackage{graphicx}
\usepackage{xcolor}
\usepackage[american]{babel}
\definecolor{cvprblue}{rgb}{0.21,0.49,0.74}
\usepackage[pagebackref,breaklinks,colorlinks,citecolor=cvprblue]{hyperref}

\def\paperID{6703} % *** Enter the Paper ID here
\def\confName{CVPR}
\def\confYear{2024}

\title{
\begin{tcolorbox}[colback=white, colframe=white, boxrule=0pt, arc=0pt, outer arc=0pt, boxsep=0pt, left=0cm, right=0cm, top=0cm, bottom=-0.5cm, width=\textwidth, enlarge left by=0cm, enlarge right by=0cm]
\centering
{\color{black}Monkey\includegraphics[width=1cm, height=1cm]{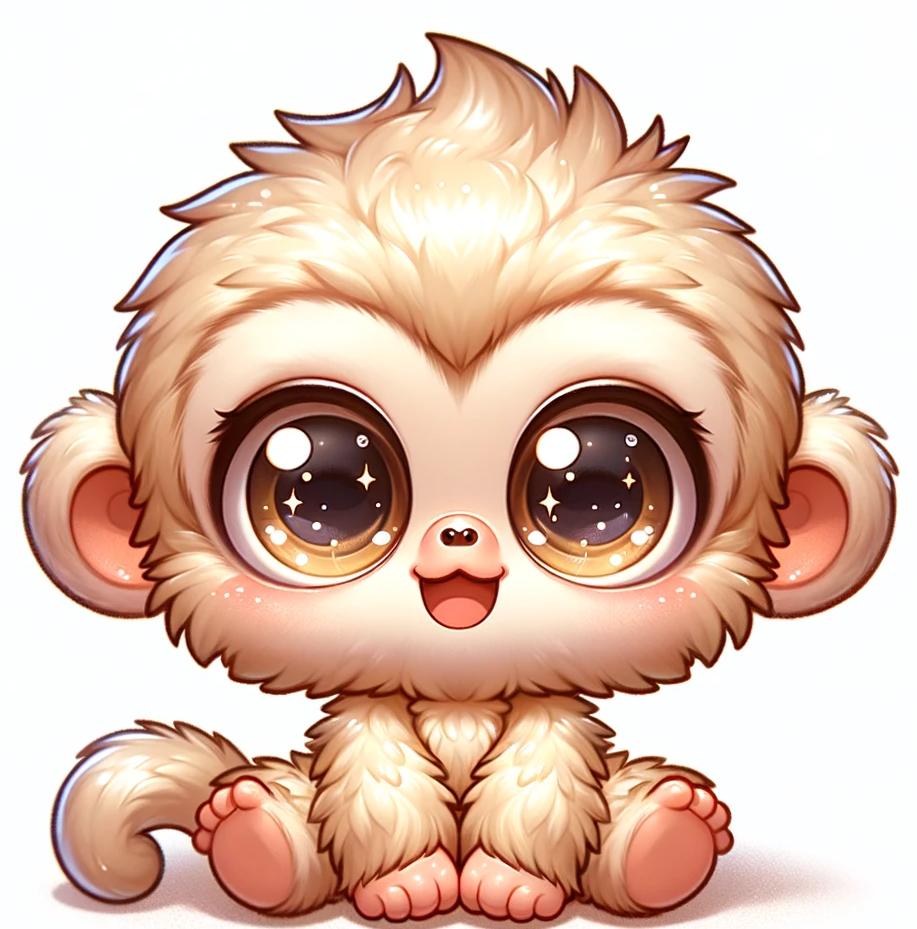}: Image Resolution and Text Label Are Important Things for Large Multi-modal Models}
\end{tcolorbox}
}
\author{%
  Zhang Li$^{1\dagger}$, Biao Yang$^{1\dagger}$, Qiang Liu$^{2}$, Zhiyin Ma$^{1}$, Shuo Zhang$^{1}$, Jingxu Yang$^{2}$, Yabo Sun$^{2}$, \\
  Yuliang Liu$^{1*}$, Xiang Bai$^{1}$\thanks{$^\dagger$equal contribution; $^*$corresponding authors}\\
  $^1$Huazhong University of Science and Technology\ $^2$Kingsoft Office\ \\
  \texttt{ylliu@hust.edu.cn} \\
}

\begin{document}
\maketitle
    \begin{abstract}
    Large Multimodal Models (LMMs) have shown promise in vision-language tasks but struggle with high-resolution input and detailed scene understanding. Addressing these challenges, we introduce Monkey to enhance LMM capabilities. Firstly, Monkey processes input images by dividing them into uniform patches, each matching the size (e.g., 448$\times$448) used in the original training of the well-trained vision encoder. Equipped with individual adapter for each patch, Monkey can handle higher resolutions up to 1344$\times$896 pixels, enabling the detailed capture of complex visual information.
    Secondly, it employs a multi-level description generation method, enriching the context for scene-object associations. This two-part strategy ensures more effective learning from generated data: the higher resolution allows for a more detailed capture of visuals, which in turn enhances the effectiveness of comprehensive descriptions. Extensive ablative results validate the effectiveness of our designs. Additionally, experiments on 18 datasets further demonstrate that Monkey surpasses existing LMMs in many tasks like Image Captioning and various Visual Question Answering formats. Specially, in qualitative tests focused on dense text question answering, Monkey has exhibited encouraging results compared with GPT4V. Code is available at \url{https://github.com/Yuliang-Liu/Monkey}.

    % \begin{figure*}[t]
    %   \centering
    %    \includegraphics[width=0.7\linewidth]{figs/abstract_new.pdf}
    %    \caption{The performance of Monkey on a broad range of multi-modal tasks compared with existing models.}
    %    \label{fig:onecol}
    % \end{figure*}
    \begin{figure}[h]
        \centering
        \includegraphics[width=1.0\linewidth]{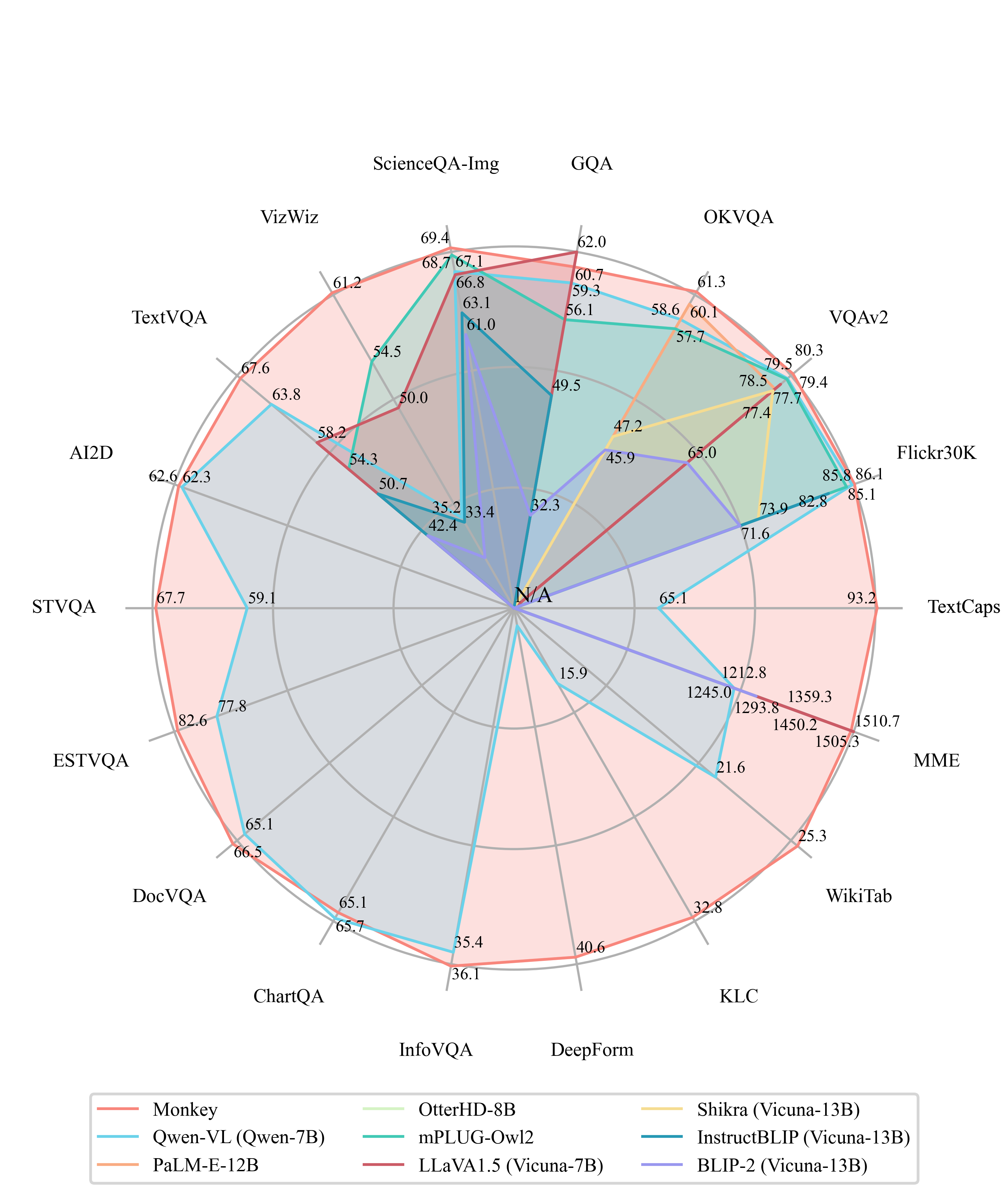}
        \caption{The performance of Monkey on a broad range of multi-modal tasks compared with existing models.}
        \label{fig:onecol}
    \end{figure}
\end{abstract}

\section{Introduction}
\label{sec:intro}
    The field of large multimodal models (LMMs) is advancing quickly because of their skill in handling different types of data, like images and text. Their success in various tasks, including image captioning and visual question answering, is attracting attention in the academic community. 

    \begin{figure*}[ht]
  \centering
    \includegraphics[width=1.0\linewidth]{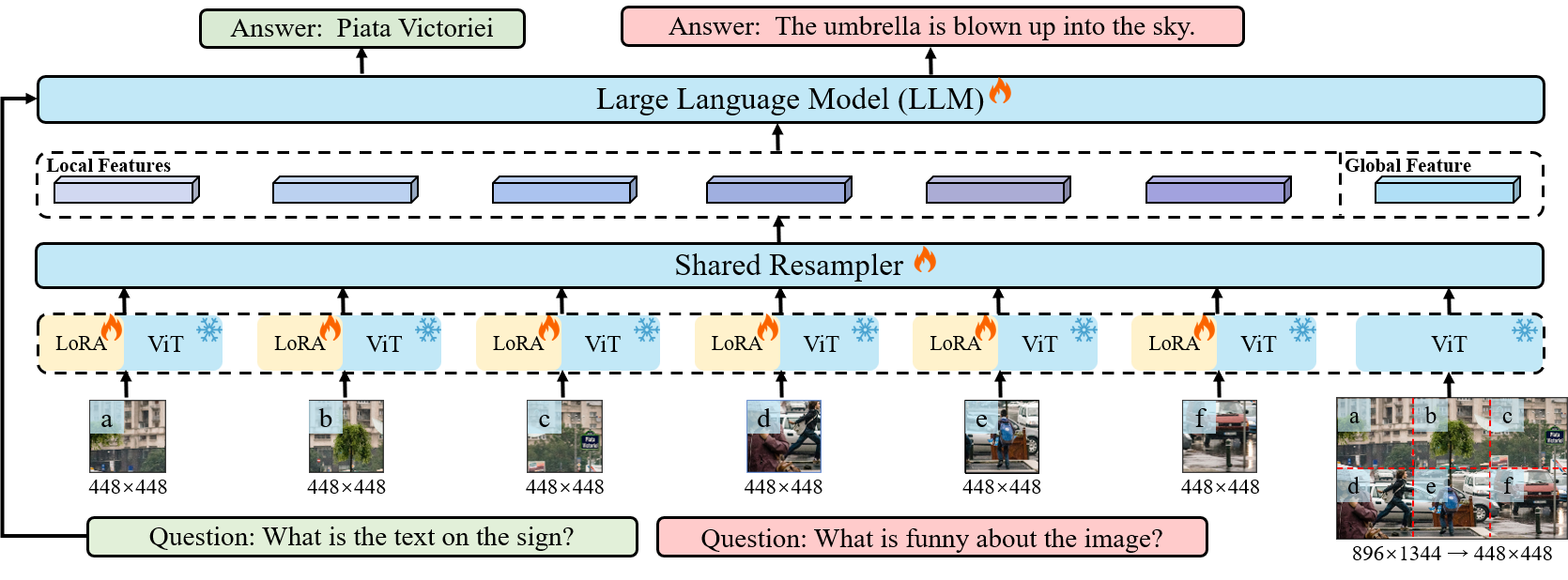}
   \caption{The overall architecture of Monkey. It enables high resolution by capturing global feature from original image and local features from divided patches. All patches are processed through the shared static Vit encoder, such as Vit-BigG with 2b parameters.}
   \label{fig:architecture}
    \end{figure*}
    
    Training LMMs benefits greatly from high-resolution images \cite{bai2023qwen-vl}, because higher resolution allows these models to detect more nuanced visual details, leading to accurate recognition of objects, their interrelationships, and the broader context within the image. 
    % The enhanced visual detail can thus contribute to generating more accurate and descriptive captions.
    Additionally, the improved visual clarity of high-resolution images aids in effectively capturing and representing complex details essential for detailed captioning. Despite advancements, handling the wide range of image resolutions and training data quality is still challenging, especially in complex situations. Solutions include using pre-trained visual modules with larger input resolution (like LLaVA1.5~\cite{liu2023llava1.5} ) and gradually increasing the resolution of the training process through curriculum learning (like Qwen-VL~\cite{bai2023qwen-vl}, PaLI-3~\cite{chen2023pali-3} and PaLI-X~\cite{chen2023pali-x}) have been explored, but they demand significant training resources and still face challenges in handling larger image sizes. To fully leverage the benefits of large input resolution, it is crucial to have more detailed image descriptions, which can enhance the understanding of image-text relationships. However, the short captions in widely used datasets such as  COYO~\cite{kakaobrain2022coyo-700m} and LAION~\cite{schuhmann2022laion} are usually intuitively insufficient.

   We introduce Monkey, a resource-efficient approach to increase input resolution within the Large Multimodal Model frameworks. 
   % Leveraging existing LMMs and avoiding extensive pre-training, Monkey utilizes a novel module that divides high-resolution images into smaller patches using a sliding window method. Each patch is processed independently by a static visual encoder, enhanced with LoRA~\cite{hu2021lora} adjustments and a trainable visual resampler. 
   Compared to the approach of directly interpolating the ViT to increase input resolution, Monkey utilizes a new module that divides high-resolution images into smaller patches using a sliding window method. Each patch is processed independently by a static visual encoder, enhanced with LoRA~\cite{hu2021lora} adjustments and a trainable visual resampler. This technique leverages existing LMMs while circumventing the need for extensive pre-training.
   % This allows for combining the patch encodings with the global image encoding, improving image understanding when presented to the language decoder. 
   The key idea is that these encoders are typically trained on smaller resolutions (like 448$\times$448), which is costly to train from scratch. By resizing each patch to its supported resolution, we maintain the training data distribution for the encoder. Our method, which uses various trainable patches to enhance resolution, shows a clear advantage over traditional interpolation techniques for positional embedding, as demonstrated by our quantitative analysis. 
    
   To further leverage the advantage of large resolution, we have also proposed an automatic multi-level description generation method. This method is designed to produce high-quality, abundant caption data by seamlessly combining insights from multiple generators. It utilizes the strengths of a diverse array of advanced systems: BLIP2~\cite{li2023blip2}, known for its nuanced image-text understanding; PPOCR~\cite{du2020pp}, a robust optical character recognition system; GRIT~\cite{wu2022grit}, which excels in granular image-text alignments; SAM~\cite{sam}, a dynamic model for semantic alignment; and ChatGPT~\cite{chatgpt}, an AI renowned for its contextual understanding and language generation capabilities. By integrating the unique capabilities of these systems, our method offers a comprehensive and layered approach to caption generation, capturing a wide spectrum of visual details.
   % This synergy of diverse technologies ensures that our captions are not only accurate and detailed but also contextually rich and varied, significantly enhancing the interpretative recognition ability of our models.

   We summarize the advantages of the Monkey as follows:
    
   \begin{enumerate}
    \item \textbf{Support resolution up to 1344$\times$896 without pretraining}.
    By going beyond the usual 448$\times$448 resolution used in LMMs, the higher resolution helps to better identify and understand small or closely grouped objects and dense text.
    
    \item \textbf{Contextual associations}. 
    We introduce a multi-level description generation method that improves the model's ability to grasp the relationships among multiple targets and more effectively utilize common knowledge in generating text descriptions. 
    % This leads to more thorough and insightful outcomes.
    
    \item \textbf{Performance enhancements on many evaluation datasets}.
    As shown in Fig.~\ref{fig:onecol}, we carried out testing across 18 diverse datasets, leading to a very competitive performance by our Monkey model in tasks such as Image Captioning, General Visual Question Answering, Scene Text-centric Visual Question Answering, and Document-oriented Visual Question Answering. 
    In particular, during qualitative evaluations centered on dense text question answering, Monkey has shown promising results, comparing with GPT4V.
   \end{enumerate}

    \begin{figure*}[t]
      \centering
       \includegraphics[width=0.95\linewidth]{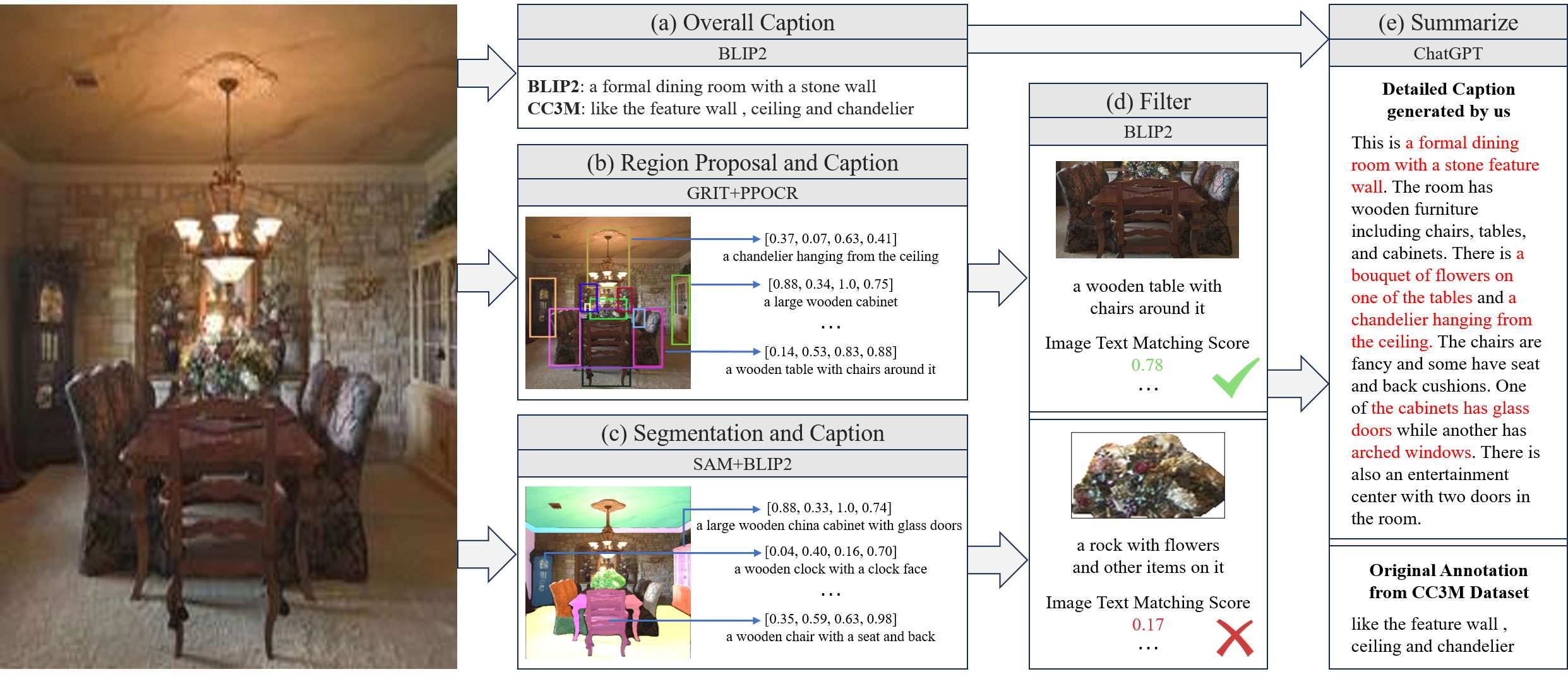}
       \caption{The pipeline for multi-level description generation for images.}
       \label{fig:generation}
    \end{figure*}

    \section{Related Work}
    \label{sec:related}
    The Large Multimodal Models (LMMs) field has seen significant progress, particularly in enhancing visual and language processing. Methods like Flamingo~\cite{alayrac2022flamingo} and OpenFlamingo~\cite{awadalla2023openflamingo} have advanced visual representation by integrating a Perceiver Resampler with vision encoders. BLIP2~\cite{li2023blip2} employs a Q-Former to link the frozen LLM and vision encoder. Unified-IO~\cite{lu2022unified} demonstrates versatility by training across over 80 diverse datasets, widening its domain applicability. PaLM-E~\cite{driess2023palm-e} adopts a unique approach by treating images and text as “multimodal sentences” to improve visual-language tasks. MiniGPT4~\cite{zhu2023minigpt4} bridges visual modules and LLMs, enhancing multimodal capabilities. InstructBLIP~\cite{dai2023instructblip}, starting from BLIP2, adds instructional inputs to the Q-Former for task-relevant visual features. MME~\cite{fu2023mme} introduces a benchmark for evaluating LMMs' perception and cognition.

    Additionally, there has been significant progress in leveraging large language models. The LLaVA series, including LLaVA~\cite{liu2023llava} and LLaVA1.5~\cite{liu2023llava1.5}, align vision encoders and LLMs for better image-text understanding. mPLUG-Owl~\cite{ye2023mplug} focuses on fine-tuning with mixed text and visual-text data. mPLUG-Owl2~\cite{ye2023mplugowl2} introduces shared modules for better modality collaboration. KOSMOS-2~\cite{peng2023kosmos2} enables visual answers like detection boxes. Shikra~\cite{chen2023shikra} specializes in Referential Dialogue, adept at processing positional inputs and outputs. BLiVA~\cite{hu2023bliva} combines task-related and global features for enhanced multimodal task processing. Qwen-VL~\cite{bai2023qwen-vl} improves visual module resolution to 448. OtterHD~\cite{li2023otterhd} fine-tunes Fuyu-8B~\cite{fuyu-8b} with instruction/response pairs, maintaining the original image size during inference.
    
    Despite these advancements, challenges remain in extracting finer image features, as noted by~\cite{liu2023hidden,xu2023lvlm}, which indicate the need for ongoing development in the field.

\section{Methods}
    Fig.~\ref{fig:architecture} illustrates the comprehensive architecture of Monkey. Initially, the input image is segmented into patches. These patches are then processed through a shared Vision Transformer (ViT) equipped with distinct adapters. Subsequently, both local and global features, along with the question, are processed using the shared resampler and the Large Language Model (LLM), resulting in the generation of the desired answers.

    \subsection{Enhancing Input Resolution}
    Input resolution is crucial for accurately interpreting text and detailed image features. Previous studies~\cite{bai2023qwen-vl,chen2023pali-3} have shown the effectiveness of starting with smaller resolutions and progressively advancing to larger ones through curriculum learning. However, this approach can be highly resource-demanding, often necessitating comprehensive pretraining with large-scale data (as seen in Qwen-VL, which supports resolutions up to 448$\times$448). 
    % Directly resizing the visual encoder through position interpolation, as detailed in Sec.~\ref{subsec:ab}, presents its own set of challenges, including limited performance and the need for increased training resources to achieve convergence. 
    To address these issues and efficiently enhance resolution, we introduce a simple yet more effective technique.
    
    Given an image $I \in \mathbb{R}^{H\times W \times 3}$, we employ a sliding window $W \in \mathbb{R}^{H_v\times W_v}$ (where $H_v, W_v$ denote the supported resolution of the original LMM) to partition the image into smaller, local sections. We also leverage LoRA~\cite{hu2021lora} within each shared encoder to address the varied visual elements in different parts of an image. This integration of LoRA is to help our encoders to recognize and assimilate detail-sensitive features from each image area effectively, which enhances the understanding of spatial and contextual relationships without a substantial increase in parameters or computational demand.

    To preserve the overall structural information of input image, we resize the original image to dimensions ($H_v, W_v$),  maintaining it as a global image. Following this, both the individual patches and the global image are processed through the visual encoder and resampler concurrently. Here, the visual resampler, inspired by Flamingo~\cite{alayrac2022flamingo}, is a mechanism that performs two main functions: summarizing visual information and obtaining higher semantic visual representations in a language feature space. It achieves this by leveraging a cross-attention module. The module employs trainable vectors (embeddings) as query vectors, along with image features from the visual encoder serving as keys for cross-attention operations. 
    
    This approach strikes a balance between detailed and holistic perspectives of the images, thereby enhancing the model performance while avoiding a substantial increase in computational demand.

    \subsection{Multi-level Description Generation}
    Previous models such as LLaVA~\cite{liu2023llava} and Qwen-VL~\cite{bai2023qwen-vl} used large datasets like LAION~\cite{schuhmann2022laion}, COYO~\cite{kakaobrain2022coyo-700m}, and CC3M~\cite{sharma-etal-2018-conceptual} for their initial training. However, these datasets often offer image-text pairs that are too simple (e.g., one short sentence to describe a complicated image), lacking in detailed imagery. As a result, even when these models are trained with high-resolution images, they struggle to accurately link visual features with basic captions. This limitation affects the models to effectively combine visual processing with language understanding.

    To bridge this gap, we develop a novel approach for generating multi-level descriptions automatically. This technique is designed to create rich and high-quality caption data by effectively blending the outputs from various generators. We utilize a combination of several advanced systems, each bringing its own strength to the process: BLIP2~\cite{li2023blip2}, which provides a deep understanding of the relationship between images and text; PPOCR~\cite{du2020pp}, a strong performer in optical character recognition; GRIT~\cite{wu2022grit}, specializing in detailed image-text matching; SAM~\cite{sam}, focused on semantic alignment; and ChatGPT~\cite{chatgpt}, known for its exceptional ability in contextual language generation. 
    
    As shown in Fig.~\ref{fig:generation}, the image description process begins with BLIP2 creating overall captions using a Q-former for tight integration with the vision encoder and LLM, while retaining original CC3M annotations for context. Next, GRIT, a region-to-text model, generates detailed descriptions of specific regions, objects, and their characteristics. PPOCR extracts text from the images, and SAM segments and identifies objects and their parts. These objects are then individually described by BLIP2. However, to counter potential inaccuracies from these tools, especially in zero-shot settings, we find it essential to further use BLIP2 to check for consistency between image areas, objects, and their descriptions, filtering out low-scoring matches. Finally, all data, including global captions, localized descriptions, text extracts, and object details with spatial coordinates, are fed into the ChatGPT API for fine-tuning, enabling ChatGPT to generate accurate and contextually rich image descriptions.

    By merging the unique features of these systems, our approach achieves a layered and comprehensive style of caption creation. It captures an extensive range of visual and textual nuances, resulting in captions that are not just elaborate, but also contextually diverse and engaging. 
    
    \begin{table}[]
    \centering
    \scalebox{0.9}{
    \begin{tabular}{c|c|c}
    \toprule 
    Task                                    & Dataset          & Smaples \\ \midrule 
    \multirow{3}{*}{Image Caption}          & Detailed Caption & 213k    \\
                                            & COCO Caption~\cite{karpathy2015coco}     & 82k     \\
                                            & TextCaps~\cite{textcaps}         & 109k    \\ \midrule 
    \multirow{5}{*}{General VQA}            & VQAV2~\cite{goyal2017making}            & 100k    \\
                                            & OKVQA ~\cite{marino2019ok}          & 18k     \\
                                            & GQA~\cite{hudson2019gqa}              & 150k    \\
                                            & ScienceQA ~\cite{lu2022learn}       & 18k     \\
                                            & VizWiz ~\cite{gurari2018vizwiz}         & 20k     \\ \midrule 
    \multirow{3}{*}{Scene Text-centric VQA} & TextVQA ~\cite{singh2019towards}         & 34k     \\
                                            & OCRVQA ~\cite{mishra2019ocr}          & 250k    \\
                                            & AI2D ~\cite{kembhavi2016diagram}            & 24k     \\ \midrule 
    \multirow{8}{*}{Doc-oriented VQA}       & DocVQA ~\cite{mathew2021docvqa}          & 118k    \\
                                            & ChartQA ~\cite{masry2022chartqa}         & 84k     \\
                                            & InfoVQA ~\cite{mathew2022infographicvqa}         & 47k     \\
                                            & DeepForm ~\cite{deepform}        & 7k      \\
                                            & KLC ~\cite{stanislawek2021kleister}             & 27k     \\
                                            & WTQ ~\cite{pasupat2015compositional}              & 28k     \\
                                            & TabFact ~\cite{chen2019tabfact}        & 91k     \\
                                            & VisualMRC ~\cite{tanaka2021visualmrc}        & 21k     \\ \midrule 
    Total                                     &  -                & 1.44m   \\ \bottomrule
    \end{tabular}}
    \caption{Details on the Monkey training data, derived entirely from publicly available datasets.}
    \label{tab:data}
    \end{table}
    
    \begin{table*}[]
    \centering
    \begin{tabular}{@{}l|cc|ccccc@{}}
    \toprule
    Model & \multicolumn{2}{c|}{Image Caption} & \multicolumn{5}{c}{General VQA} \\
    & Flickr30K & TextCaps & VQAv2 & OKVQA & GQA & ScienceQA &VizWiz \\ \midrule
    Flamingo-80B ~\cite{alayrac2022flamingo}       &  67.2      & -    & 56.3 & 50.6 & -    & -    & 31.6 \\
    % Unified-IO-XI~\cite{lu2022unified}           &-  & -    & 77.9 & 54.0 & -    & -    & -    \\
    % Kosmos-2 ~\cite{peng2023kosmos2}               &80.5   & -    & 51.1 & -    & -    & -    & -    \\
    Palm-E-12B ~\cite{driess2023palm-e}             &-  & -    & 77.7 & \uline{60.1} & -    & -    & - \\
    BLIP-2 (Vicuna-13B) ~\cite{li2023blip2}     &71.6 & -    & 65.0 & 45.9 & 32.3 & 61.0 & 19.6 \\
    InstructBLIP  (Vicuna-13B) ~\cite{dai2023instructblip}        &82.8   & -    & -    & -    & 49.5 & 63.1 & 33.4 \\
    Shikra (Vicuna-13B) ~\cite{chen2023shikra}   &73.9    & -    & 77.4 & 47.2& -    & -    & -    \\
    mPLUG-Owl2 ~\cite{ye2023mplugowl2}      &  85.1       & -  & 79.4 & 57.7 &56.1  & \uline{68.7} & \uline{54.5}       \\
    LLaVA1.5 (Vicuna-7B) ~\cite{liu2023llava1.5}  &-    & -    & 78.5& -& \textbf{62.0}    & 66.8    & 50.0    \\
    Qwen-VL(Qwen-7B) ~\cite{bai2023qwen-vl}    &  \uline{85.8}   & \uline{65.1}    & \uline{79.5} & 58.6 & 59.3 & 67.1 & 35.2 \\
    Qwen-VL-Chat  ~\cite{bai2023qwen-vl}      &81.0       &- & 78.2 & 56.6 & 57.5 & 68.2 & 38.9 \\ \midrule
    Monkey   &\textbf{86.1}     & \textbf{93.2}    & \textbf{80.3} & \textbf{61.3} & \uline{60.7} & \textbf{69.4} & \textbf{61.2} \\ \bottomrule
    \end{tabular}
    \caption{Results on Image Caption and General VQA.}
    \label{General VQA}
    \end{table*}
    
    \begin{table}[]
    \centering
    \footnotesize
    \begin{tabular}{@{}l|cccc@{}}
    \toprule
    Model & TextVQA  & AI2D & STVQA & ESTVQA \\ \midrule 
    Pix2Struct-Large ~\cite{lee2023pix2struct}   & -             & 42.1          & -             & -             \\
    BLIP-2~\cite{li2023blip2}          & 42.4          & -             & -             & -             \\
    InstructBLIP~\cite{dai2023instructblip}        & 50.7          & -             & -             & -             \\
    mPLUG-DocOwl ~\cite{ye2023mplug}        & 52.6          & -             & -             & -             \\
    mPLUG-Owl2~\cite{ye2023mplugowl2}          & 54.3          & -             & -             & -             \\
    Qwen-VL~\cite{bai2023qwen-vl}            & \uline{63.8}  & \uline{62.3}          & \uline{59.1}  & \uline{77.8}  \\ % TextVQA为paper结果，其他为ori结果
    Qwen-VL-Chat~\cite{bai2023qwen-vl}         & 61.5          & 57.7  & -             & -             \\
    LLaVA-1.5~\cite{liu2023llava1.5}         & 58.2          & -             & -             & -             \\ \midrule
    Monkey  & \textbf{67.6}    & \textbf{62.6}    & \textbf{67.7}    & \textbf{82.6}    \\ \bottomrule
    \end{tabular}
    \caption{Results on Scene Text-centric VQA.}
    \label{TextVQA}
    \end{table}
    
    \begin{table}[h]
    \centering
    \footnotesize
    \scalebox{0.9}{
    \begin{tabular}{@{}l|cccccc@{}}
    \toprule
    Model & DocVQA & ChartQA & InfoVQA & DeepForm & KLC & WTQ\\ \midrule
    Qwen-VL             & 65.1          & \textbf{65.7}  & 35.4          & 4.1           & 15.9          & 21.6  \\  % DocVQA和ChartQA为paper结果，其他为ori结果 
    Monkey  & \textbf{66.5}  & 65.1          & \textbf{36.1}          & \textbf{40.6}  & \textbf{32.8}    & \textbf{25.3}    \\ \bottomrule
       \end{tabular}}
    \caption{Results on Doc-oriented VQA.}
    \label{DocVQA}
    \end{table}

    \begin{table*}[]
    \centering
    
    \scalebox{0.85}{
    \begin{tabular}{@{}c|c|c|cc|cccccccc@{}}
    \toprule
    &Resolution  & LoRA &Throughout & FLOPS (e20)
 & VQAv2       & GQA         & TextVQA        & STVQA          & DocVQA & DeepForm   & InfoVQA & WTQ     \\ \midrule
    r1 & 896$\times$896*     & 0  &43.452& 1.608  &74.1 &55.2 & 44.7 & 41.5 &53.9 &11.4 &32.7 &16.8        \\
    r2 & 896$\times$896*    & 1  &37.429& 1.614   &  71.4  & 54.0 & 41.7 & 38.5 & 47.5 &7.2 &31.5 &17.1        \\ \midrule
    r3 &672$\times$672     & 4  &43.604& 1.617   & 80.0          & 59.6        & 67.3           & \uline{67.2}   & 66.4   & 31.3 &35.9 &25.0        \\
    r4 &784$\times$784     & 4  &42.851& 1.617   & 79.9        & 59.8        & 67.5   & \textbf{67.7}  & \uline{66.5}      & 38.9 &35.5 &25.1 \\
    r5 &896$\times$1344     & 6 &28.542& 1.622     &80.1        & \uline{61.1}        &67.3   &66.7  &66.3      &\textbf{42.3} &\textbf{39.6} &\textbf{26.6} \\
    r6  &1344$\times$896   & 6 &28.842& 1.622 & \uline{80.2} & \textbf{61.8} & \textbf{67.7} & 66.3 & 64.5 & \uline{41.4} & 35.7 & 25.2 \\    \midrule
    r7 &896$\times$896     & 0 &49.634& 1.613
    & 80.1   &  60.4       & 67.5          & 65.1      & 66.1   & 36.8 &36.1 &24.9         \\
    r8 &896$\times$896     & 1 &42.885 & 1.614   & 80.0             &  60.3        & \uline{67.6}          & 67.0      & \textbf{66.7}   & 36.9 &\uline{36.5} &24.7         \\
    r9 &896$\times$896     & 4 &42.542 &  1.617
   & \textbf{80.3}     & 60.7    & \uline{67.6}    & \textbf{67.7}  & \uline{66.5}      & 40.6  &36.1 &\uline{25.3}  \\ \bottomrule
    \end{tabular}}
    \caption{Ablation study on enhancing input resolution and the number of trainable adapters using Qwen-VL (originally trained using 448$\times$448). * refers to directly scaling the input size of the visual encoder from 448 to 896 using traditional positional position interpolation. }
    \label{SizeAblation}
    \end{table*}

    \subsection{Multi-task Training}
    Our goal is to train a model that is both cost-effective and capable of understanding different types of images for various tasks. By integrating various datasets and employing uniform instructions for all tasks, as guided by~\cite{bai2023qwen-vl}, we enhance the model's learning ability and training efficiency. 
    % This unified approach ensures that the model gains a comprehensive understanding of different image types, enabling it to perform effectively across various tasks.

   We focus on tasks such as creating image captions, responding to image-based questions, and other activities requiring the model to process both text and images. For captioning, we instruct the model with ``Generate the caption in English:'' for basic captions, and ``Generate the detailed caption in English:'' for more intricate ones. When it comes to answering questions about images, we use a straightforward format: ``\{question\} Answer: \{answer\}.'' 
  
    In our training process, we use a variety of public datasets tailored to specific tasks. For image captioning, we include both our own detailed captions and established datasets like COCO caption~\cite{karpathy2015coco} and TextCaps~\cite{textcaps}. For general Visual Question Answering (VQA), we utilize datasets such as VQAV2~\cite{goyal2017making}, OKVQA~\cite{marino2019ok}, GQA~\cite{hudson2019gqa}, ScienceQA~\cite{lu2022learn}, and VizWiz~\cite{gurari2018vizwiz}. For Text-centric VQA tasks, we select TextVQA~\cite{singh2019towards}, OCRVQA~\cite{mishra2019ocr}, and AI2D~\cite{kembhavi2016diagram}; while for document-related VQA, we employ datasets like DocVQA~\cite{mathew2021docvqa}, ChartQA~\cite{masry2022chartqa}, InfoVQA~\cite{mathew2022infographicvqa}, DeepForm~\cite{deepform}, Kleister Charity (KLC)~\cite{stanislawek2021kleister}, WikiTableQuestions (WTQ)~\cite{pasupat2015compositional}, TableFact~\cite{chen2019tabfact}, and VisualMRC~\cite{tanaka2021visualmrc}. We use our multi-level description generation method to regenerate around 427k image-text pairs from the CC3M dataset, previously used in LLaVA's pretraining phase. To ensure balanced training, we control the image count for each task as detailed in Tab.~\ref{tab:data}. Our compiled dataset, with around 1.44 million examples, is designed to train our model effectively in understanding and executing various instructions. 

    \section{Experiment}
    We evaluate our model by testing it across a spectrum of standard vision-language tasks, including the generation of image descriptions, answering diverse visual questions, and comprehending targeted phrases in images. 
    
    \subsection{Implementation Details}
    \textbf{Model Configuration.}
    We conduct experiments based on the well-trained Vit-BigG~\cite{ilharco_gabriel_2021_5143773} and LLM from Qwen-VL~\cite{bai2023qwen-vl}, the pre-trained large multimodal model. Since the vision encoder has already been well pretrained, we proceed directly to the instruction-tuning stage. During instruction tuning, $H_v$, $W_v$ are set to 448 to match the encoder of Qwen-VL. 
    % The number of learnable queries of the visual resampler is 256. 
    We employ a consistent resampler across all crops. The learnable queries engage with local features, utilizing the same set of 256 learnable queries for each crop.
    Due to limitations in training time, our main experiments were mainly conducted using images of size 896$\times$896 unless specify. For LoRA, we set the rank to 16 for the attention module and 32 for MLP in the encoder. Monkey includes 7.7B parameters for a large language model, with 90M parameters for the resampling module, an encoder with 1.9B parameters, and 117M parameters for LoRA. The overall parameters for Monkey is 9.8B.

    \textbf{Training.}
     During the training process, we utilize the AdamW optimizer~\cite{adamw} with a learning rate of 1e-5 and the cosine learning rate schedule. Additionally, we set the values of $\beta_1$ and $\beta_2$ to 0.9 and 0.95, respectively. We incorporate a warmup period of 100 steps and employ a batch size of 1024. To control overfitting, we apply a weight decay of 0.1. The whole training process takes 40 A800 days for one epoch.

    \subsection{Results}
    We report the results on Image Caption, General VQA, Scene Text-centric VQA, and Document-oriented VQA. We also conduct testing on the MME benchmark and achieve a perception score of 1505.3, ranking second, as shown in Fig.~\ref{fig:onecol}. The details of each dataset can be found in Appendix.
    
    \textbf{Image Caption.}
    Image captioning is vital for connecting visual content with the understanding of natural language. In our study, we select Flickr30K~\cite{young2014image} and TextCaps~\cite{textcaps} as the benchmark for testing the image captioning task.  TextCaps challenges the model to interpret and reason text within images effectively. We present our model's performance on Flickr30K and TextCaps in Tab.~\ref{General VQA}, where the results indicate that Monkey demonstrates enhanced performance on these datasets. We also qualitatively show effectiveness of our method in offering detailed image descriptions in Sec.~\ref{subsec:vis} and Appendix.

    \textbf{General VQA.} 
    General visual question answering (VQA) requires ability to learn visual and textual information, showing a deep understanding of how they interrelate. For General VQA, we validate on five benchmarks: VQAv2~\cite{goyal2017making}, OKVQA~\cite{marino2019ok}, GQA~\cite{hudson2019gqa}, ScienceQA~\cite{lu2022learn}, and VizWiz~\cite{gurari2018vizwiz}. The performance results are shown in Tab.~\ref{General VQA}. Our model shows remarkable proficiency in VQAV2, OKVQA, ScienceQA, and VizViz, surpassing the nearest competing method by an average of 1.62\%. These results highlight the effectiveness of our method, emphasizing its use of high input resolution and detailed data.

    \textbf{Scene Text-centric VQA.} Text information is commonly found in real-world scenes, making the ability to answer questions about text in images a crucial aspect of question-answering tasks. For our evaluation, we employ four datasets: TextVQA~\cite{singh2019towards}, AI2D~\cite{kembhavi2016diagram}, STVQA~\cite{STVQA}, and ESTVQA~\cite{ESTVQA}. The results, shown in Tab.~\ref{TextVQA}, indicate that our model leads in performance on these datasets, outperforming the nearest competitor by an average of 4.37\%. Based on our observation, this enhanced performance is mainly attributed to the increased image resolution, which brings smaller text and finer details into clearer view. Moreover, the inclusion of detailed caption data during training provides valuable textual context, further boosting the robustness of the model.

    \begin{figure*}[t!]
      \centering
       \includegraphics[width=1.0\linewidth]{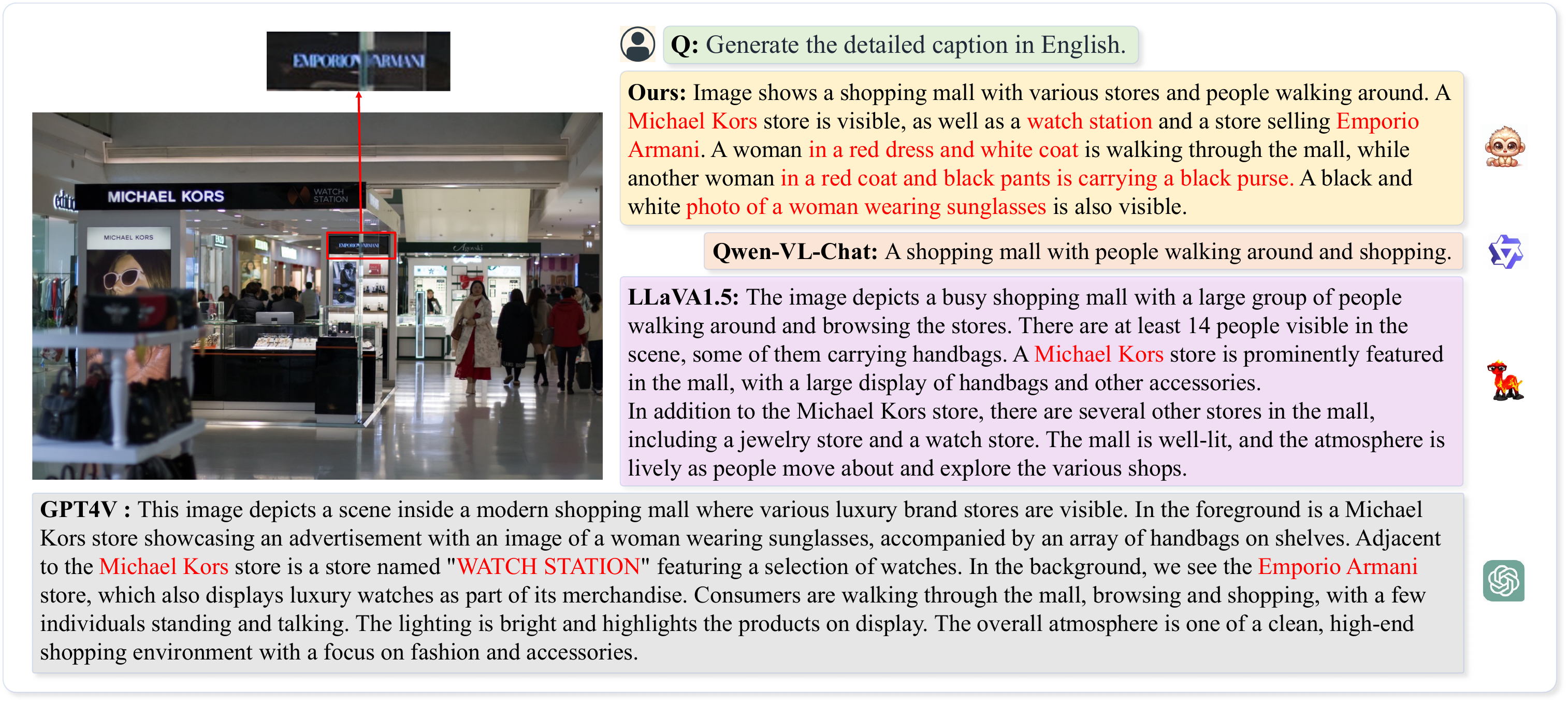}
       \caption{Visualization comparisons with existing LMMs on Detailed Caption task. Accurate and specific descriptions ar marked in red. More examples refer to Appendix.}
       \label{Densecap_vs_GPT4V}
    \end{figure*}  
    
    \textbf{Document-oriented VQA.} 
    Despite the clean backgrounds of documents, their densely packed text poses distinct challenges. To effectively evaluate our model, we select representative benchmarks including DocVQA~\cite{mathew2021docvqa}, ChartQA~\cite{masry2022chartqa}, InfographicVQA~\cite{mathew2022infographicvqa}, DeepForm~\cite{deepform}, KLC~\cite{stanislawek2021kleister}, and WTQ~\cite{pasupat2015compositional}. The results, as detailed in Tab.~\ref{DocVQA}, show that Monkey surpasses Qwen-VL in most Document-oriented VQA tasks, achieving an averagely significant improvement of 9.77\%. The higher resolution of documents reveals more intricate details and a denser concentration of information. Monkey's capability to process larger input resolutions enhances its spatial perception, thereby improving its recognition and comprehension of various document elements like text, charts, infographics, and forms.

    % \section{Analysis}
    % \label{sec:analysis}
    
    \subsection{Ablation Study}
    \label{subsec:ab}
    We conduct thorough experiments to validate the effectiveness of our designs.  
    
    \textbf{Ablation study on strategies of enhancing input resolution.} 
    We first evaluate the existing technique of improving input resolution, as illustrated in Tab.~\ref{SizeAblation}. Resizing the visual encoder using traditional positional position interpolation
    to a size of 896 results in worse performance compared with our method under the same settings (r1 vs. r9). 
    Interestingly, applying LoRA to the encoder for this traditional interpolation method appears to be less effective than not using it (r1 vs. r2). This may due to the inherited parameters from the previous encoder are specifically tuned by lower resolution, changing it by force may necessitate more training resources. 
    
    % For our method (r3-r9), as we increase the input size, there is a noticeable boost in performance, especially demonstrated in the DeepForm dataset. The model's ability to discern intricate details and sharper images enhances its understanding of visual aspects such as objects, shapes, and textures, thereby improving its overall visual perception. When we further push the input resolution to 1344$\times$896, which is the highest resolution the device can support, the model shows further improvements on high-resolution datasets like DeepForm, InfoVQA, and WTQ, as detailed in Tab.~\ref{SizeAblation}. However, we can note that for some datasets, such as TextVQA, using the largest resolution results in a slight decline in performance; however, the original average resolution in the TextVQA dataset is around 950 pixels in width and 811 pixels in height, further increasing its input resolution seems unnecessary for these images.
    For our method (r3-r9), as we increase the input size, there is a noticeable boost in performance, especially demonstrated in the DeepForm dataset. It can be observed that adding LORA does not significantly increase FLOPs and the use of one LORA or four LORAs results in a minimal difference in throughput (r7-r9). The model's ability to discern intricate details and sharper images enhances its understanding of visual aspects such as objects, shapes, and textures, thereby improving its overall visual perception. When we further push the input resolution to 1344$\times$896, which is the highest resolution the device can support, the model shows further improvements on high-resolution datasets like DeepForm, InfoVQA, and WTQ, as detailed in Tab.~\ref{SizeAblation}. However, we can note that for some datasets, such as TextVQA, using the largest resolution results in a slight decline in performance; nevertheless, the original average resolution in the TextVQA dataset is around 950 pixels in width and 811 pixels in height, further increasing its input resolution seems unnecessary for these images. 

    Furthermore, as shown in Tab.~\ref{Tab:llava15_ablation}, we consistently demonstrate the effectiveness of our method on LLaVA1.5. Impressively, we noticed significant improvements when we increased the input resolution from 224 to 448, demonstrating the efficiency of our approach. 
    
    \begin{figure*}[t!]
    \centering
    \includegraphics[width=0.96\linewidth]{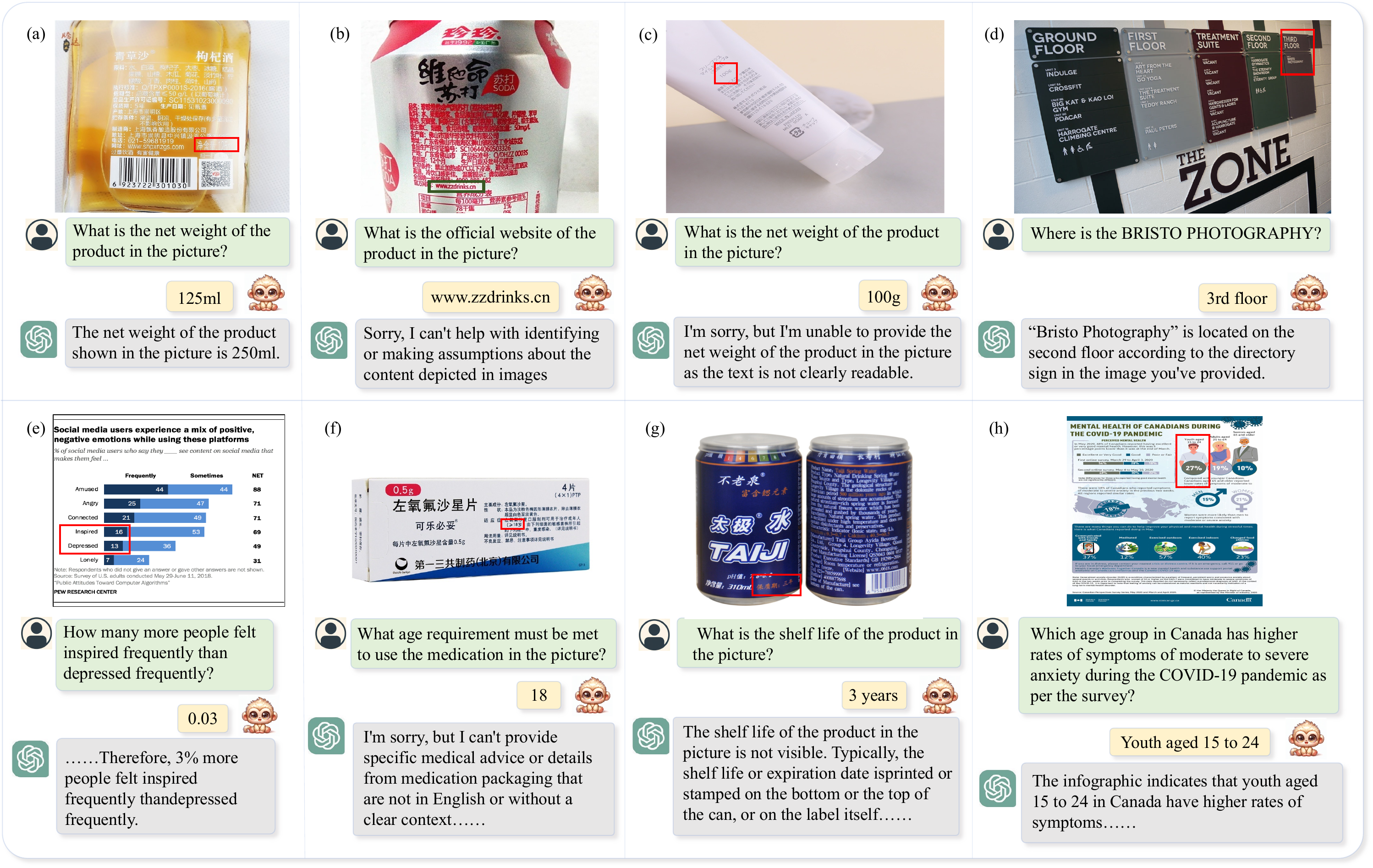}
    \caption{On some testing samples with dense text, Monkey has demonstrated impressive performance when compared to GPT4V.
    % The testing samples are strictly checked to guarantee they are not from training samples.
    }
    \label{Doc_Chart}
    \end{figure*}
    
    \textbf{Trainable Adapters.}  As shown in Tab.~\ref{SizeAblation}, reducing the LoRA number causes a performance decrease. Using one LoRA for all patches compared to not using LoRA provides a better perception of local details (r7 vs. r8), especially with a significant improvement in STVQA. Utilizing four LoRA modules leads to a better performance, which may because this approach enables the model to learn a better understanding of the spatial relationships and contextual information within distinct image regions.

\begin{table}[h]
\centering
\scalebox{0.95}{

\begin{tabular}{cc|cccc}
 \toprule
\multicolumn{1}{l}{Res.} & PT   & \multicolumn{1}{c}{GQA} & \multicolumn{1}{c}{TextVQA} &  \multicolumn{1}{c}{MMVet} \\  \midrule
 224                      & CC3M & 62                      & 56.1                                     & 33.2                      \\
224                      & Ours & 62.1(+0.1)                  & 56.3(+0.2)             & 33.7(+0.5)                 \\
\midrule
336                      & CC3M & 63.4                    & 59.8                     & 33.5                      \\
336                      & Ours & 63.7 \textcolor{red}{\textbf{(+0.3)}}                   & 60.4(+0.6)         & 36.1 \textcolor{red}{\textbf{(+2.6)}}                     \\  \midrule

448                      & CC3M & 64.3                    & 60.2                     & 33.6                      \\
448                      & Ours & 64.6 \textcolor{red}{\textbf{(+0.3)}}                   & 62.0 \textcolor{red}{\textbf{(+1.8)}}          & 36.2 \textcolor{red}{\textbf{(+2.6)}}                     \\ 
\bottomrule
\end{tabular}}
\caption{Ablation study on LLaVA1.5. ``Res.'' denotes resolution. ``PT'' refers to pretrain data.}
\label{Tab:llava15_ablation}
\end{table}

    % \textbf{Multi-level Description.}
    % In order to assess the impact of our generated data, we carried out ablation studies on LLaVA1.5. We employed a 336-resolution ViT-L as our vision encoder and Vicuna13B~\cite{vicuna2023} as the language model. Our experiments involved a comparative analysis where we pretrained LLaVA1.5 
    % using the same original data from LLaVA, with the modification of replacing 427k text with our generated annotations. 
    % The instruction tuning data were exclusively sourced from LLaVA1.5. Our multi-level generated data show notable improvements in performance across three VQA tasks – GQA, VizWiz, and TextVQA – and on three key evaluation benchmarks: POPE~\cite{POPE}, MMBench~\cite{liu2023mmbench}, and MMVet~\cite{yu2023mm}, as detailed in Tab. ~\ref{Tab:llava15_ablation}. We attribute these enhancements to the detailed captions provided during pretraining, which enables the model to better focus on various objects and their attributes in images, thereby improving the coordination between the vision module and the language model.

    \textbf{Collaboration between High Resolution and Multi-level Description.} To validate the collaboration between High Resolution and Multi-level Description, we conduct ablation studies on LLaVA1.5. We employ a ViT-L as our vision encoder and Vicuna13B~\cite{vicuna2023} as the language model. By replacing the original annotation from CC3M with our generated annotations in the pretraining, we consistently achieved better results on GQA, TextVQA and MMVet~\cite{yu2023mm}, as demonstrated in Tab.~\ref{Tab:llava15_ablation}. Furthermore, we have observed that detailed descriptions consistently yield greater performance enhancements at resolutions of 336 and 448, compared to a resolution of 224. In Appendix, we provide visualization results for Monkey at different resolutions. These results show that models with high resolution shines when trained with more comprehensive descriptions.

    \subsection{Visualization}
    \label{subsec:vis}
    In a side-by-side qualitative analysis, we compared Monkey with GPT4V and other LMMs on a task of generating detailed captions. The results, illustrated in Fig.~\ref{Densecap_vs_GPT4V}, demonstrate Monkey's superior capability in providing exhaustive descriptions of images. For instance, in the image from Fig.~\ref{Densecap_vs_GPT4V}, both Monkey and GPT4V successfully identified an ``Emporio Armani'' store in the background. Moreover, Monkey went further in detailing various elements in the scene,
    such as describing 
    % a woman ``in a red dress and white coat'', 
    ``another woman in a red coat and black pants carrying a black purse''.
    % , and a billboard featuring ``a woman wearing sunglasses'' on the image's left side. 
    % More visualization results of Monkey can be found in Appendix.
    
    Additionally, as shown in Fig.~\ref{Doc_Chart}, we qualitatively observe that in many cases for understanding complex text-based inquiries, Monkey has shown impressive performance when compared to GPT4V. More visualization results of Monkey can be found in Appendix.
   
   % These examples from the visual question answering (VQA) task illustrate various strengths of our model. 
   % In (a), the model not only shows its relevance to the task at hand but also its ability to utilize world knowledge effectively. Example (b) demonstrates Monkey's precision in pinpointing answers, even in areas with dense and indistinct text. Lastly, (c) demonstrates the effectiveness of our model in accurately recognizing a wide array of objects. Additionally, the enhanced input resolution of our model significantly boosted its performance with document images, especially those laden with dense text. Example (d) illustrates our model's skill in extracting exact answers from high-resolution and complex posters. Example (e) highlights the model's adeptness in logical reasoning and computation using chart images. Lastly, (f) displays the model's ability to accurately retrieve information from document images with complicated layouts.

\subsection{Limitation}
    The capability of our method to process input images is constrained to a maximum of six patches due to the limited input length of the language model. This restriction hampers the further expansion of input resolution. 

    Moreover, for the multi-level description generation approach, it is capable of describing only the scene presented in the image and its scope is bound by the world knowledge encapsulated in BLIP2 and the original CC3M annotations. For instance, when provided with a photo of a location in a country, the method can describe the visual aspects of the scene, but it lacks the ability to identify and specify that the scene is indeed in which country.
    % or pinpoint any other exact location. On the other hand, while we have generated 427k data instances, this quantity is relatively modest for Large Multimodal Models (LMMs) and incurred a cost of about 500 dollars. Therefore, it is worthwhile to investigate more sophisticated pipelines and consider expanding the volume of generated data.

\section{Conclusion}
    This paper proposes a training-efficient approach to effectively improve the input resolution capacity up to 1344$\times$896 pixels without pretraining from the start. To bridge the gap between simple text labels and high input resolution, we propose a multi-level description generation method, which automatically provides rich information that can guide the model to learn the contextual association between scenes and objects. With the synergy of these two designs, our model achieved excellent results on multiple benchmarks. By comparing our model with various LMMs, including GPT4V, our model demonstrates promising performance in image captioning by paying attention to textual information and capturing fine details within the images; its improved input resolution also enables remarkable performance in document images with dense text. 
\section*{Acknowledgements}
This research is supported by NSFC (No.\  62225603, No.\  62206104).
    % In conclusion, this paper further highlights that image resolution and text labels are crucial for the effectiveness of Large Multi-modal Models, providing valuable insights for future research in this domain.
    
   % 0_abstract}  
% \input{sec/1_intro}
% \input{sec/2_related}
% \input{sec/3_methods}
% % \input{sec/4_experiment}
% \input{sec/5_analysis}
% \input{sec/6_conclusion}
{
    \small
    \normalem
    \bibliographystyle{ieeenat_fullname}
    \bibliography{main}
}

\clearpage
\onecolumn
\appendix

\section{Summary of the Evaluation Benchmarks.}
We present a comprehensive overview of the evaluation benchmarks utilized, along with their corresponding metrics in Tab. ~\ref{tab:benchmark}. For the Image Caption task, we selected two datasets: Flickr30K~\cite{young2014image}, which is an image caption dataset for natural images, and TextCaps~\cite{textcaps}, which is an image caption dataset for natural images with text. For general Visual Question Answering (VQA), we chose five commonly used datasets. VQAV2~\cite{goyal2017making} is an open-ended VQA dataset focused on natural images, while OKVQA~\cite{marino2019ok} requires additional world knowledge. GQA~\cite{hudson2019gqa} is a dataset designed for real-world visual reasoning and compositional question answering. ScienceQA~\cite{lu2022learn} involves multimodal multiple-choice VQA on science topics, and VizWiz~\cite{gurari2018vizwiz} aims to answer questions from blind individuals. In the domain of Scene Text-centric VQA, our selection includes TextVQA~\cite{singh2019towards}, AI2Diagram~\cite{kembhavi2016diagram}, STVQA~\cite{STVQA}, and ESTVQA~\cite{ESTVQA}. AI2D is a multiple-choice VQA dataset that focuses on science diagrams, while the others involve reading and reasoning about text in natural images. For the STVQA and ESTVQA datasets, we followed the split provided by ~\cite{liu2023hidden}. Regarding Doc-oriented VQA, we encompass various document images, including documents, charts, infographics, reports, and HTML tables. In the case of DeepForm~\cite{deepform} and KLC~\cite{stanislawek2021kleister}, we transform the Key Information Extraction task into a Visual Question Answering (VQA) task. Additionally, we evaluate Monkey on the MME benchmark~\cite{fu2023mme}, which measures perception and cognition abilities. Furthermore, for the ablation study on LLaVA1.5 ~\cite{liu2023llava1.5}, we adhere to the evaluation settings specified by LLaVA1.5.

\begin{table*}[!h]
    \centering
    \scalebox{0.75}{
    \begin{tabular}{l|l|l|l|l}
        \toprule
         Task & Dataset & Description & Split & Metric  \\
         \midrule
         \multirow{2}{*}{Image Caption}
         & Flickr30K~\cite{young2014image} & Image caption for natural images   & karpathy-test & CIDEr($\uparrow$) \\
         & TextCaps~\cite{textcaps} & Image caption for natural images with text & val & CIDEr($\uparrow$) \\
         \midrule
         \multirow{5}{*}{General VQA} & VQAv2~\cite{goyal2017making} & Open-ended VQA about natural images & val & VQA Score($\uparrow$) \\
         & OKVQA~\cite{marino2019ok} &  VQA involving world knowledge on natural images & val & VQA Score($\uparrow$) \\
         & GQA~\cite{hudson2019gqa} & Real-world visual reasoning and compositional question answering & test-dev & Accuracy($\uparrow$) \\
         & ScienceQA~\cite{lu2022learn}  & Multimodal multiple choice VQA on science topics & test & Accuracy($\uparrow$) \\
         & VizWiz~\cite{gurari2018vizwiz} & Answering visual questions from blind people & val & VQA Score($\uparrow$)\\
         \midrule
         \multirow{4}{*}{Scene Text-centric VQA}
         & TextVQA~\cite{singh2019towards} & VQA involving reading and reasoning about text & val & VQA Score($\uparrow$) \\
         & AI2Diagram~\cite{kembhavi2016diagram} & Multiple choice VQA on science diagrams & test & Accuracy($\uparrow$) \\
         & STVQA~\cite{STVQA} & VQA involving reading and reasoning about text & test* & ANLS($\uparrow$) \\
         & ESTVQA~\cite{ESTVQA} & VQA involving reading and reasoning about text & test(English)* & ANLS($\uparrow$) \\
         \midrule
         \multirow{6}{*}{Doc-oriented VQA}
         & DocVQA~\cite{mathew2021docvqa} &  VQA on document images & test & ANLS($\uparrow$) \\
         & ChartQA~\cite{masry2022chartqa} & VQA on charts with visual and logical reasoning & test & Relaxed Accuracy($\uparrow$) \\
         & InfoVQA~\cite{mathew2022infographicvqa} & VQA on infographic images & test & ANLS($\uparrow$) \\
         & DeepForm~\cite{deepform} & Key Information Extraction on charity organizations‘ reports & test & Accuracy($\uparrow$) \\
         & KLC~\cite{stanislawek2021kleister} & Key Information Extraction on documents related to election spending  & test & Accuracy($\uparrow$) \\
         & WTQ~\cite{pasupat2015compositional} & VQA on semi-structured HTML tables sourced from Wikipedia & test & Accuracy($\uparrow$) \\
         \midrule
         \multirow{1}{*}{Evaluation Benchmark} 
          & MME~\cite{fu2023mme} & Evaluation benchmark measuring perception and cognition abilities  & Perception & Accuracy ($\uparrow$)\\
         \bottomrule
    \end{tabular}}
    \caption{Summary of the evaluation benchmarks.}
    \label{tab:benchmark}
\end{table*}

\newpage
\section{More Visualization Results}
\begin{figure*}[!h]
  \centering
   \includegraphics[width=0.95\linewidth]{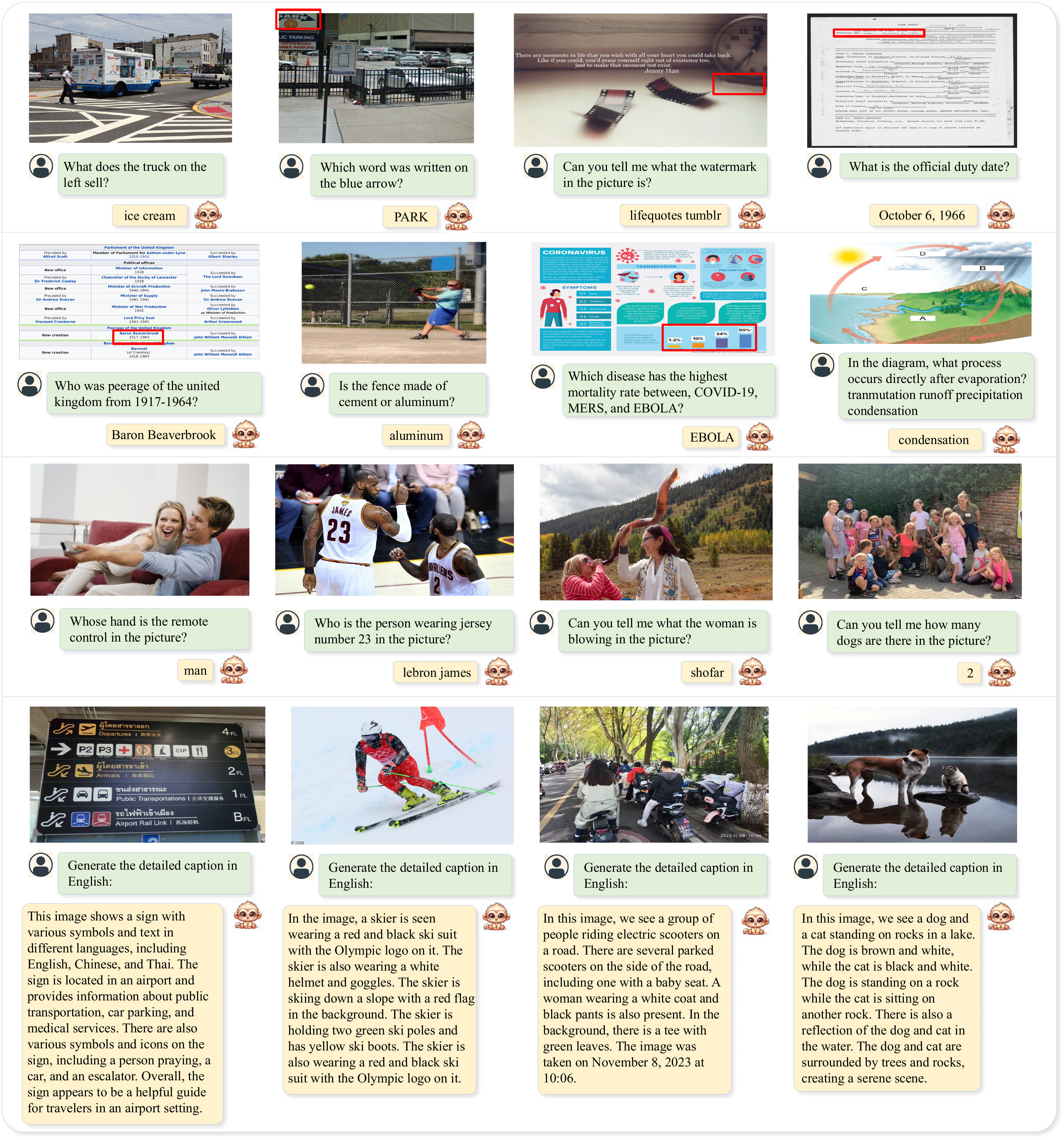}
   \caption{Visualization results.}
   \label{QA_ability}
\end{figure*} 

We presented additional visualization results, where Fig.~\ref{QA_ability} demonstrates Monkey's capabilities in various VQA tasks. Monkey analyzes the question, identifies the key elements in the image relevant to answering the question, and exhibits the ability to perceive even minute text within the image. Moreover, Monkey can reason about the objects present in the scene and possesses a strong understanding of visual charts. In addition, Fig. \ref{QA_ability} also showcases Monkey's impressive captioning ability, accurately describing various objects in the image and providing appropriate summaries.

\newpage
\section{More Examples of our Generated Data }
\begin{figure*}[h!]
  \centering
   \includegraphics[width=0.95\linewidth]{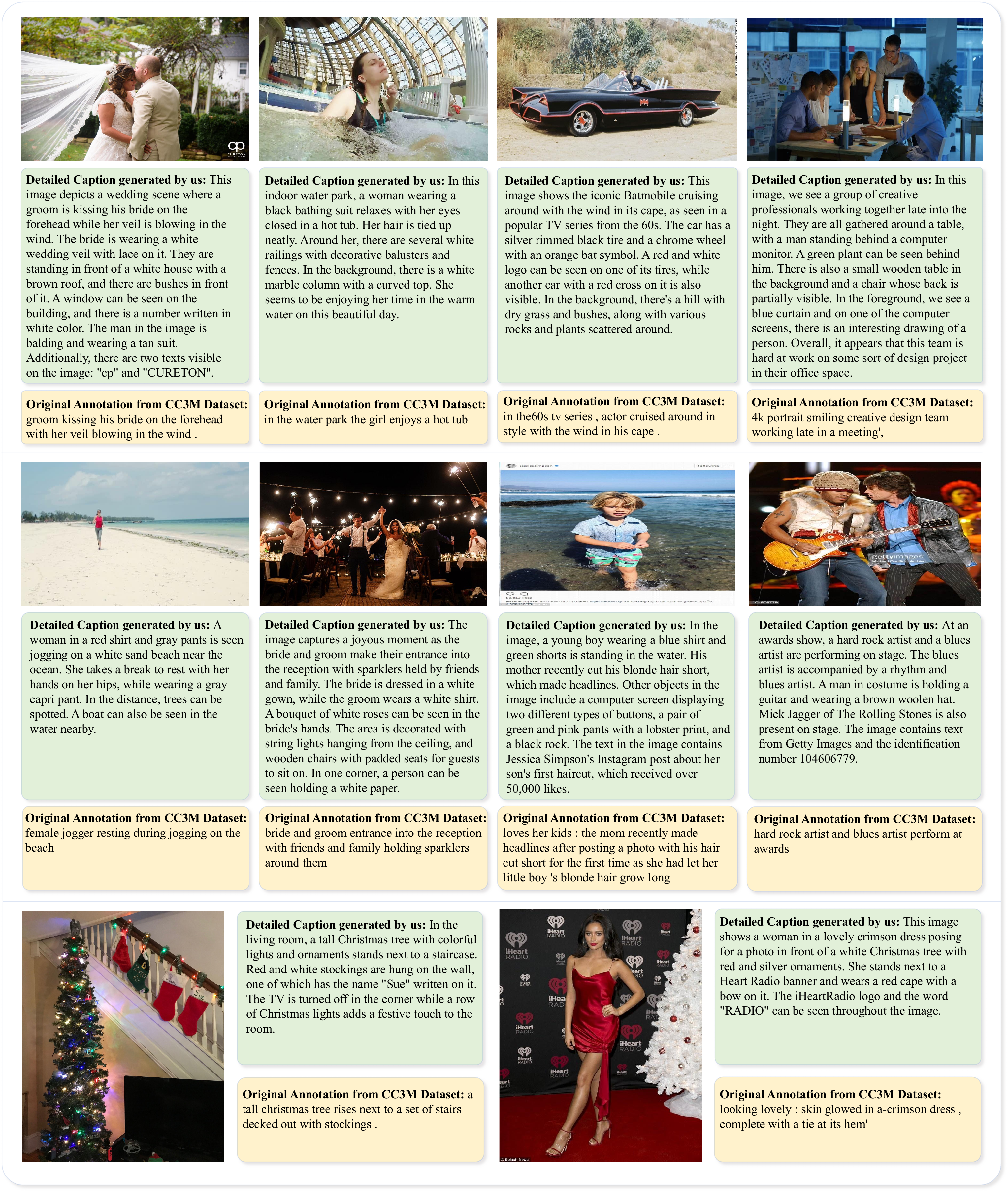}
   \caption{Detailed captions generated by us.}
   \label{dense_text}
\end{figure*} 
In Fig. \ref{dense_text}, we present the detailed captions generated by our method. Compared to the original annotations from the CC3M ~\cite{sharma-etal-2018-conceptual}, our generated descriptions cover many more details of the image, providing a more detailed description of the image.

\section{Comparison with other LMMs.}
\begin{figure*}[!h]
  \centering
   \includegraphics[width=0.95\linewidth]{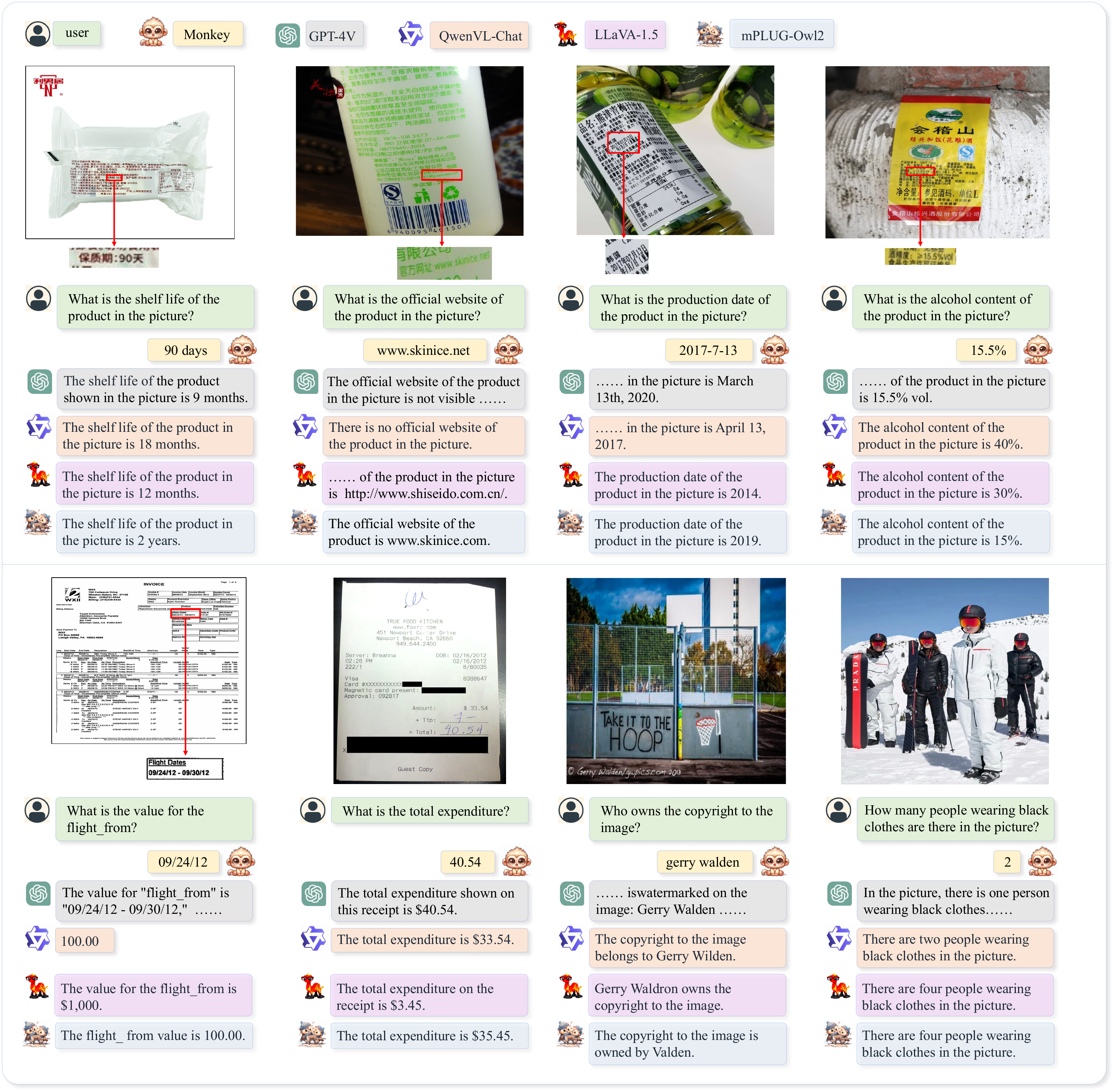}
   \caption{Visualization of Monkey's comparison with GPT-4V, QwenVL-Chat, LLaVA-1.5, and mPLUG-Owl2 on VQA task.}
   \label{QA_compare}
\end{figure*}

\begin{figure*}[!h]
  \centering
   \includegraphics[width=0.95\linewidth]{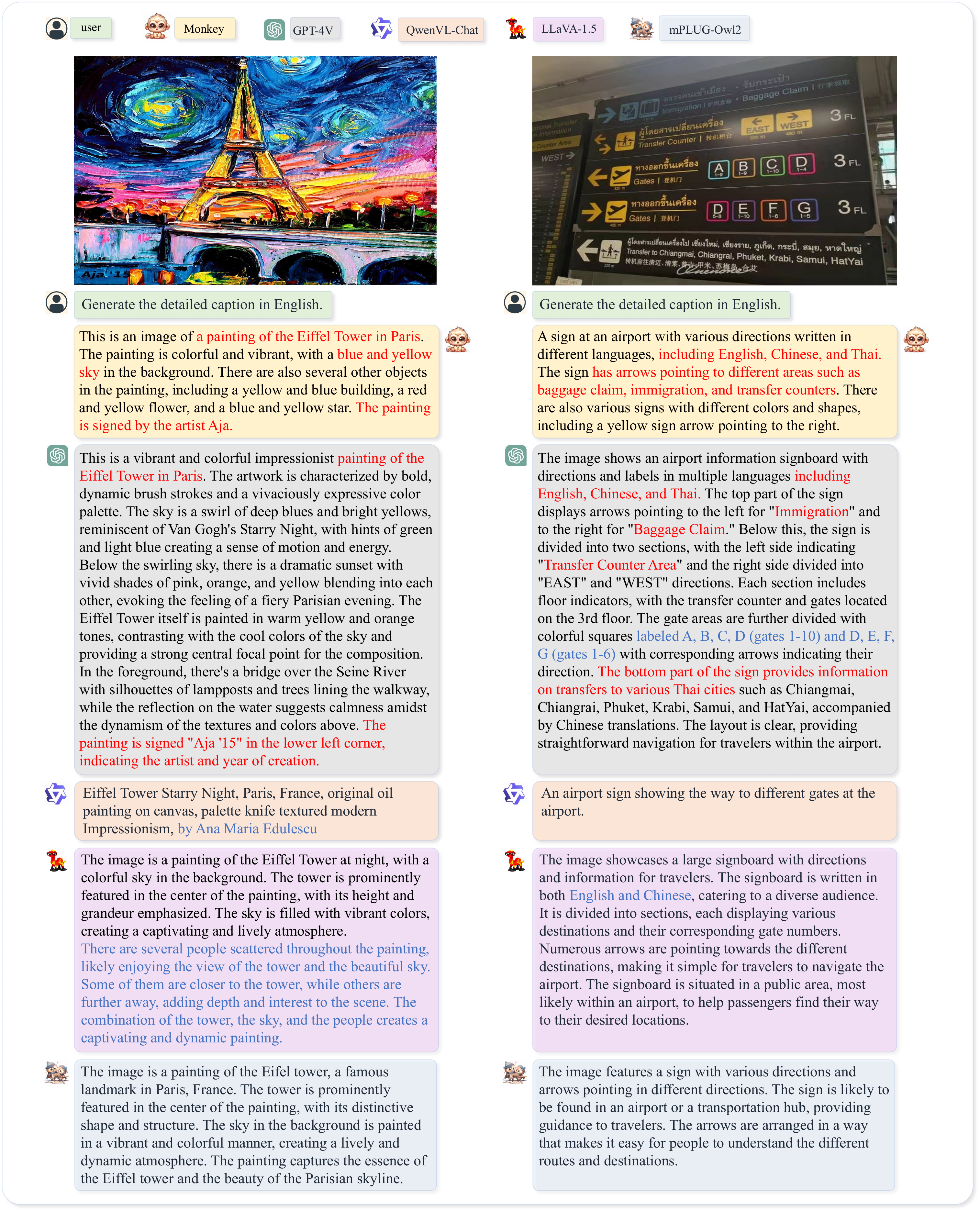}
   \caption{Visualization of Monkey's comparison with GPT-4V, QwenVL-Chat, LLaVA-1.5, and mPLUG-Owl2 on Detailed Caption task. Accurately described text is marked in red, while inaccurately described text is marked in blue.}
   \label{Caption_compare}
\end{figure*}
The comparison results of the VQA task in Fig. \ref{QA_compare} indicate that after applying our method of scaling up the model size, Monkey has achieved significant performance advantages in tasks related to dense text. It not only surpasses the performance of QwenVL-Chat ~\cite{bai2023qwen-vl}, LLaVA-1.5 ~\cite{liu2023llava1.5}, and mPLUG-Owl2 ~\cite{ye2023mplugowl2} but also achieves promising results compared to GPT-4V ~\cite{openai2023gpt4} in tasks related to dense text. This clearly demonstrates the importance of scaling up the model size for performance improvement in multimodal large models. It further validates the effectiveness of our method in enhancing the performance of multimodal large models.

In Fig. \ref{Caption_compare}, the comparison between Monkey and GPT-4V, QwenVL-Chat, LLaVA-1.5, and mPLUG-Owl2 on Detailed Caption task is shown. It can be observed that Monkey accurately describes the content of the image and exhibits high sensitivity to the text within the image. It provides detailed descriptions of the image while ensuring accuracy.
\newpage
\section{Visualization results for models at different resolutions.}
\begin{figure*}[h!]
  \centering
   \includegraphics[width=0.95\linewidth]{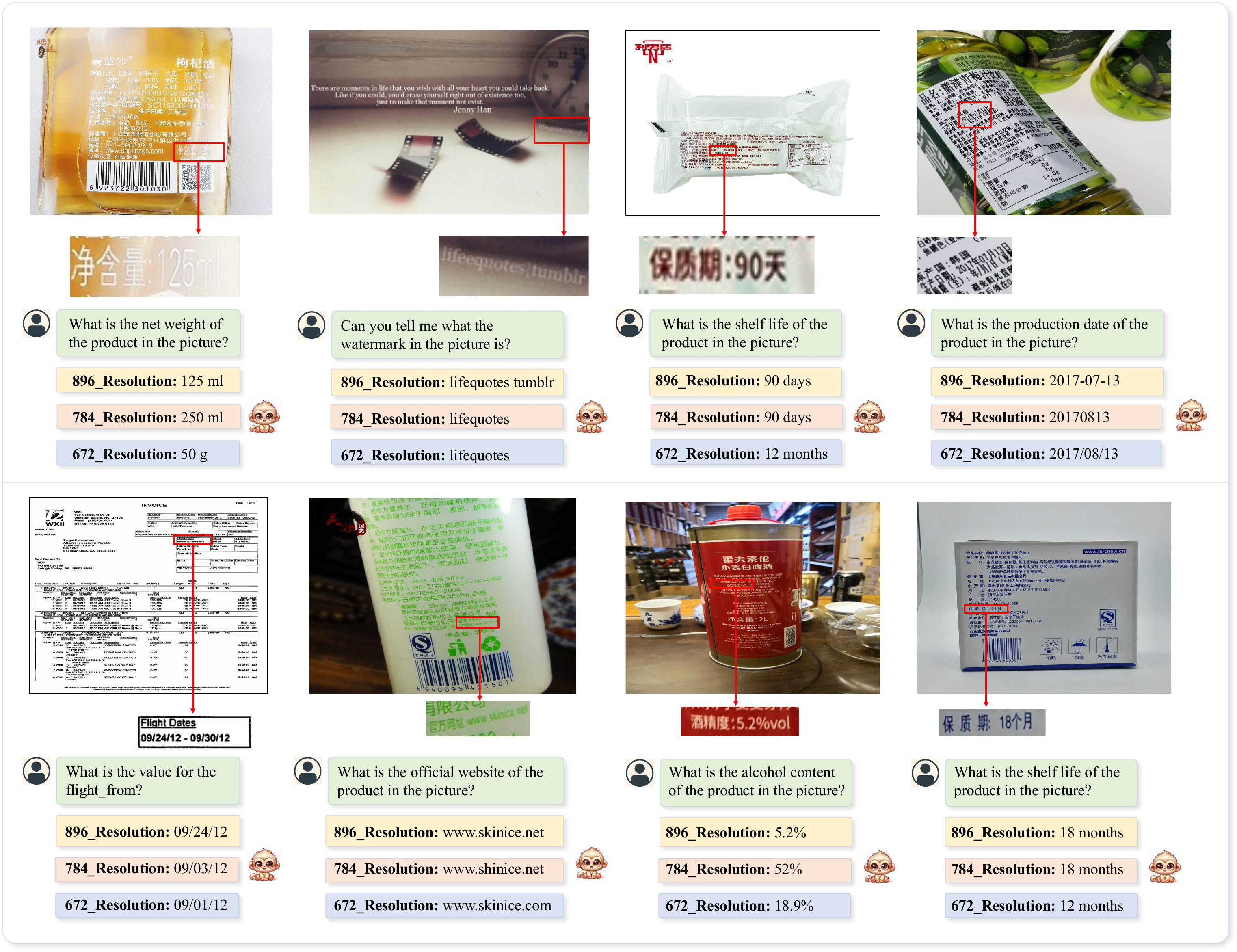}
   \caption{Visualization results of the VQA task at resolutions of 896, 784, and 672 respectively.}
   \label{QA_res}
\end{figure*} 
In Fig. \ref{QA_res}, we performed VQA tasks testing at three different resolutions: 896, 784, and 672. The visual results obtained further validate the importance of our size expansion method for improving the performance of LMMs. While using a resolution of 896 for VQA tasks testing yielded correct results, using resolutions of 784 and 672 resulted in errors, with the smallest size of 672 showing more errors.

\begin{figure*}[!t]
  \centering
   \includegraphics[width=0.95\linewidth]{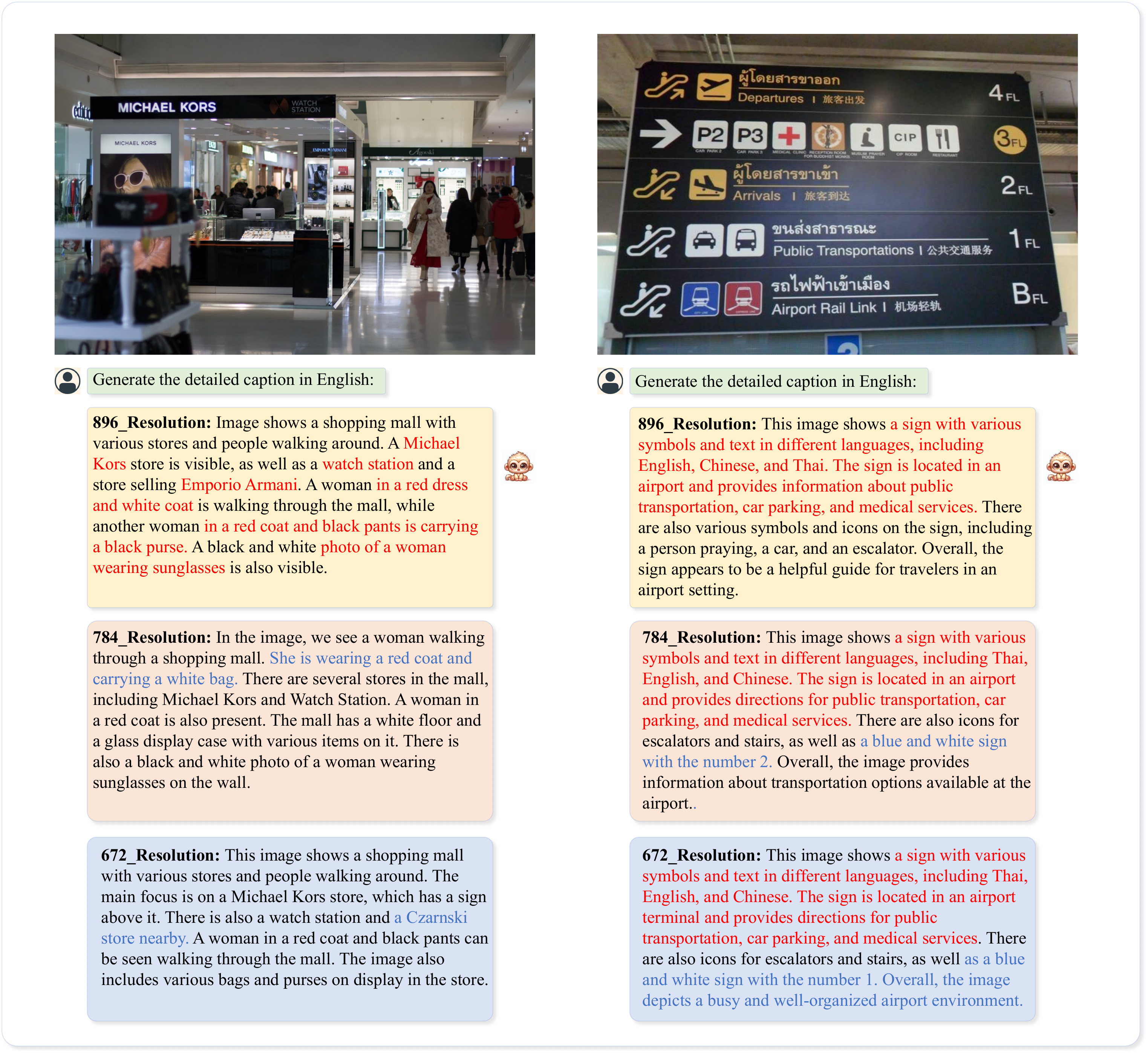}
   \caption{Visualization results of the detailed caption task at resolutions of 896, 784, and 672 respectively. Accurately described text is marked in red, while inaccurately described text is marked in blue.}
   \label{Caption_res}
\end{figure*} 
In Fig. \ref{Caption_res}, we conducted tests at three different resolutions: 896, 784, and 672. It can be observed that as the resolution decreases, the details in the images gradually become less visible to the model.

\newpage
\section{Data Generation.}
\textbf{Hyperparameter Control in Data Generation Pipeline.} The appropriate selection of hyperparameters is crucial. We empirically selected them based on qualitative results, finding SAM's default threshold and a 0.5 Image-Text Matching Score to be effective. We conducted a quantitative validation on 80 samples using the GPT-4V evaluation. The results shown in Tab.~\ref{Tab:hyper} reveal that SAM's threshold is relatively robust, and the 0.5 threshold for Image-Text Matching Score offers a better performance. 

\begin{table}[h]
\centering
\scalebox{1}{
\begin{tabular}{c|ccc}
\toprule
Pred-IOU-Thresh of SAM & 0.4    & 0.6   & 0.88 (default)  \\
GPT-4V Score     & 6.388 & 6.425 & 6.625 \\ \midrule
Image-Text Matching Score  & 0.2    & 0.5   & 0.7   \\
GPT-4V Score     & 5.825  & 6.625 & 6.550  \\ \bottomrule
\end{tabular}}
\caption{Hyperparameter Control.}
\label{Tab:hyper}
\end{table}

\newpage
\textbf{Comparison with LLaVA's GPT4 Method.} 
While the GPT4 method in LLaVA is dependent on using manually annotated captions from the COCO dataset as a foundational basis for data generation, our approach focuses on generating original, detailed captions autonomously. 
Additionally, our detectors are skilled in revealing a spectrum of details in images, from text to nuanced object characteristics, which enables to enrich unlabeled data by extracting complex, multi-level details, paving the way for the creation of both cost-effective and accurate image descriptions.

\textbf{Why choose BLIP2?} We found that the performance is very similar in the GPT-4V evaluation when utilizing brief descriptions of local areas from other VLMs, as shown in Tab.~\ref{Tab:othervlm}. However, for generating approximately 5M descriptions, using BLIP2 takes approximately 3 days, while LLaVA and mPLUG-Owl require about 21 days and 32 days, respectively. For the sake of saving time, we choose BLIP2.
\begin{table}[h]
\centering
\scalebox{1}{
\begin{tabular}{c|ccc}
\toprule
Model & LLaVA    & mPLUG-Owl   & Blip2  \\ \midrule
GPT-4V Score     & 6.663  & 6.225 & 6.625 \\ \bottomrule
\end{tabular}}
\caption{Performance of Different LMM.}
\label{Tab:othervlm}
\end{table}

\section{Ablation study on Global Feature.} We conducted experiments on the presence or absence of global features at a resolution of 896. By adding global features, the results showed a 7.5\% performance gain on TextVQA, a 0.6\% performance gain on GQA, and a 6.2\% performance gain on DocVQA. This demonstrated that global features contribute to enhancing the overall performance.


\end{document}